%% file: 2020_Nguyen_Goulet_DL.tex
\documentclass[twoside,11pt]{article}
\title{Analytically Tractable Inference in\\ Deep Neural Networks}
\author{ \textsc{Luong Ha~Nguyen}\footnote{{Correspondence: luong-ha.nguyen@polymtl.ca, james.goulet@polymtl.ca} }  ~~and \textsc{James-A.~Goulet}\footnotemark[1]\ 
\mbox{}\\ 
\small Department of Civil, Geologic and Mining Engineering\\
\small \textsc{Polytechnique Montréal}, \underline{CANADA}\\}
\date{\today}

\usepackage{hyperref}       
\usepackage{url}            
\usepackage{booktabs}       
\usepackage{amsfonts,amssymb,latexsym,amscd,mathtools}       
\usepackage{nicefrac}       
\usepackage{microtype}      
\usepackage{bm}
\usepackage{multirow}
\usepackage{caption}
\usepackage{subfig}
\usepackage{graphicx}
\usepackage{todonotes}
\usepackage{soul}

\usepackage{cancel}

\begin{document}

\maketitle

\begin{abstract}
Since its inception, deep learning has been overwhelmingly reliant on backpropagation and gradient-based optimization algorithms in order to learn weight and bias parameter values. The \emph{Tractable Approximate Gaussian Inference} (TAGI) algorithm was shown to be a viable and scalable alternative to backpropagation for shallow fully-connected neural networks. In this paper, we are demonstrating how TAGI matches or exceeds the performance of backpropagation, for training classic deep neural network architectures.  Although TAGI's computational efficiency is still below that of deterministic approaches relying on backpropagation, it outperforms them on classification tasks and matches their performance for information maximizing generative adversarial networks while using smaller architectures trained with fewer epochs.

\end{abstract}

\section{Introduction}\label{S:INTRO}
Since its inception, deep learning has been overwhelmingly reliant on backpropagation and gradient-based optimization algorithms in order to learn weight and bias parameter values. The \emph{Tractable Approximate Gaussian Inference} (TAGI) algorithm \cite{goulet2020tractable} was shown to be a viable and scalable alternative to backpropagation. Nevertheless, the method's performance was so far only demonstrated for shallow fully-connected neural networks, because its formulation is not compatible with existing platforms, e.g., TensorFlow or Pytorch. In this paper, we are demonstrating how TAGI matches or exceeds the performance of backpropagation, for classic deep neural network architectures. Section \ref{S:TAGI} first presents the fundamentals of TAGI on which section \ref{S:DAF} builds in order to adapt the approach to convolutional neural networks (CNN) \cite{lecun1995convolutional} and generative adversarial networks (GAN) \cite{goodfellow2014generative}. In parallel, this same section presents how common tools such as hidden state normalization \cite{goodfellow2016deep}, i.e. batch normalization \cite{ioffe2015batch} and layer normalization \cite{ba2016layer}, and observation noise decay \cite{DBLP:journals/corr/NeelakantanVLSK15} can be integrated into the TAGI's framework. 
Finally, section \ref{S:Benchmarks} compares the performance of TAGI with the state-of-the-art (SOTA) performance for image classification and image generation benchmarks.

\section{TAGI}\label{S:TAGI}
TAGI \cite{goulet2020tractable} relies upon two key concepts. First, the joint distribution between the observations and a neural network's parameters is approximated by a multivariate Gaussian distribution. Learning the parameters, i.e., the mean vector and covariance matrix defining the joint probability density function (PDF) requires two simplifying assumptions; (1) the product of Gaussian random variables is also Gaussian with its moments, expected values, covariance, and cross covariances calculated analytically using moment generating functions, and (2) the non-linear activations such as rectified linear unit (ReLU), Tanh, and logistic sigmoid are locally linearized at the expected value of the input hidden unit, thus  the output expected value, covariance, and cross-covariance for the activation unit can be computed analytically. The second key concept employed by TAGI has two folds; (1) it requires simplifying the covariance matrices for hidden layers and parameters to a diagonal structure, and (2) it performs the Gaussian conditional inference using a recursive layer-wise procedure which allows for an analytically tractable inference that scales linearly with respect to the number of weight parameters. Despite the simplifying assumptions mentioned above, TAGI was shown to match the performance of feedforward neural networks (FNN) trained with backpropagation for regression and classification benchmarks. In the following section, we will show how to apply TAGI to deep architecture formulations in order to demonstrate its superiority on benchmark problems.

\section{Deep architecture formulations}\label{S:DAF}
The fundamental mathematical operations, i.e., additions, multiplications and non-linear activations, in deep neural networks are no different from those in FNN. Therefore, the mathematical formulation developed for defining the joint PDF for observations and parameters in FNN can be readily employed in deeper architectures. This section covers the specificities related to CNNs and GANs.
\subsection{CNN}\label{SS:CNN}
The main building blocks of CNNs are the convolution and pooling layers whereas the convolution operation is performed using fixed-size filters whose weights are learned from data. As the convolution operation \cite{he2015convolutional} involves the same operations as a FNN, i.e., multiplication, addition and non-linear transformation, TAGI's mathematical developments can be directly applied to perform the operations in the convolution layer. Small changes occur in the pooling operations which reduce the output size following the convolutional layers. Additional changes also take place in two key tools that are employed with CNNs; hidden states' normalization, and observation noise decay. 
\subsubsection{Pooling layer}
Existing methods such as the stochastic max pooling introduced by Peters and Welling \cite{peters2018probabilistic} can be directly be applied with TAGI. In addition to this method, we adapt an \emph{average} pooling approach to the context of TAGI, where, as its name indicates, the output is the average of the activation units in the pooling kernel. For a $\mathtt{K}$-elements pooling kernel, the output Gaussian random variable is defined as
\begin{equation}
A^{\text{pool}} = \tfrac{1}{\mathtt{K}}(A_{1}+A_{2}+\cdots+A_{\mathtt{K}}), 
\end{equation}

\subsubsection{Layer and batch normalization}
In the context of TAGI, \emph{layer normalization} \cite{ba2016layer} can be done using the following principles; For a layer of Gaussian activation units $\bm{A}=[A_{1}~A_{2}\cdots A_{\mathtt{A}}]^{\intercal}$, the normalization is defined as

\begin{equation}
\label{eq:norm}
\tilde{\bm{A}}=\frac{\bm{A}^{(j)} - {\mu}_{\bm{A}}}{{\sigma}_{\bm{A}}},
\end{equation}
where ${\mu}_{\bm{A}}$ and ${\sigma}_{\bm{A}}$ are the mean and standard deviation of a Gaussian mixture reduction \cite{runnalls2007kullback} made from all activation units, each having a probability $1/\mathtt{A}$ so that 
\begin{equation}
\begin{array}{rcl}
{\mu}_{\bm{A}}&=& \tfrac{1}{\mathtt{A}}\sum_{i=1}^{\mathtt{A}} \mu_{A_{i}} \\[6pt]
{\sigma}_{\bm{A}}&=& \sqrt{\tfrac{1}{\mathtt{A}}\left[\sum_{i=1}^{\mathtt{A}}\left(\sigma_{A_{i}}\right)^{2} +\sum_{i=1}^{\mathtt{A}}\left(\mu_{A_{i}}-{\mu}_{\bm{A}}\right)^{2}\right]}.
\label{EQ:LN_1}
\end{array}
\end{equation}
The resulting expected value, covariance, and cross-covariance for the normalized vector of activation units $\tilde{\bm{A}}$ are given by
\begin{equation}
\begin{array}{rcl}
\bm{\mu}_{\tilde{\bm{A}}}&=& \frac{\bm{\mu}_{{\bm{A}}}-\mu_{\bm{A}}}{\sigma_{\bm{A}}}\\[6pt]
\bm{\Sigma}_{\tilde{\bm{A}}}&=& \frac{1}{\sigma_{\bm{A}}^2}\bm{\Sigma}_{{\bm{A}}} \\[6pt]
\text{cov}(\tilde{\bm{A}},\bm{A})&=& \frac{1}{\sigma_{\bm{A}}}\bm{\Sigma}_{{\bm{A}}}.
\end{array}
\label{EQ:norm_act}
\end{equation}
%
%
By using the short-hand notation defined by Goulet et al.~\cite{goulet2020tractable} for hidden units at a layer $j+1$, i.e. $\bm{Z}^{\text{+}}\equiv \bm{Z}^{(j+1)}$, and for the weights, bias, an hidden units at a layer $j$, i.e. $\{\bm{W},\bm{B}, \bm{Z}\}\equiv\{\bm{W}^{(j)}, \bm{B}^{(j)}, \bm{Z}^{(j)}\}$, we can use the terms from equation \ref{EQ:norm_act} in the TAGI's formulation to update the covariance between hidden states at successive layers 
\begin{equation}
\label{eq:hiddenStateNorm}
\text{cov}(\bm{Z}, \bm{Z}^{+})=\tfrac{1}{{\sigma}_{\bm A}}\bm{\Sigma}_{\bm{Z}}\mathbf{J}\bm\mu_{\bm W},
\end{equation}
where $\mathbf{J}$ is the diagonal Jacobian matrix of the activation function evaluated at $\bm{\mu}_{\bm{Z}}$, i.e.,
$\mathbf{J}=\text{diag}\big(\nabla_{\!\!\bm{z}} {\sigma}(\bm{\mu}_{\bm{Z}})\big)$. 
The covariance between the weight and bias parameters at a layer $j$ and the hidden states at a layer $j+1$ are 
\begin{equation}
\label{eq:paramNorm}
\begin{array}{rcl}
\text{cov}(\bm{W}, \bm{Z}^{+}) &=& \tfrac{1}{{\sigma}_{\bm{A}}}\bm{\Sigma}_{\bm{W}}\mu_{\bm{A}} -\tfrac{{\mu}_{\bm{A}}}{{\sigma}_{\bm{A}}}\bm{\Sigma}_{\bm{W}}\\[6pt]
\text{cov}(\bm{B},  \bm{Z}^{+}) &=& \bm{\Sigma}_{\bm{B}}.
\end{array}
\end{equation}
The developments of Equations \ref{eq:hiddenStateNorm} and \ref{eq:paramNorm} are provided in Appendix \ref{Appendix:A}. These updated expected value, covariance, and cross-covariance terms can be directly employed in the TAGI's layer-wise recursive inference procedure. 

The procedure for \emph{batch normalization} \cite{ioffe2015batch} is analogous to the layer normalization except that the normalization is made for each hidden units, over a batch of $\mathtt{B}$ observations. The Gaussian activation units contributing to the normalization are  $\bm{A}_j=[A_{j}^1~A_{j}^2\cdots A_j^{\mathtt{B}}]^{\intercal}$, so that the number of activation units $\mathtt{A}$ in Equations \ref{EQ:LN_1} and \ref{EQ:norm_act} is replaced by the number of observations per batch $\mathtt{B}$.

\subsubsection{Observation noise decay}
In the original TAGI formulation for FNN, the observation errors' standard deviation parameters $\sigma_{V}$ were considered as constants during the training. Throughout empirical experimentation with CNNs, we noticed that we could improve the performance when using a decay equation
\begin{equation}
\label{eq:noiseDecay}
\sigma_{V}^{e} = \eta\cdot\sigma_{V}^{e-1},
\end{equation}
where $e$ is the current epoch and $\eta\in(0, 1]$ is the decay factor hyperparameter that needs to be learned outside of the TAGI analytic inference procedure.
This approach is similar to what is done in standard deep neural networks trained with backpropagation where noise is added to the gradient  \cite{DBLP:journals/corr/NeelakantanVLSK15} with a decay schedule. In the case of gradient-based learning, this noise consists in discrete samples added to the gradient itself whereas for TAGI, it consists in additional variance on the output layer so that the update during the inference step will put more weight on the prior rather than on the likelihood, with this effect diminishing with time.


\subsection{Generative adversarial networks}
\label{ss:gan}

\begin{figure}[b!]
\input{tikz_setup_GAN.tex}
\begin{center}
\subfloat[Discriminator]{\scalebox{0.9}{\begin{tikzpicture}
\node[input](I)[label=center: Noise]{};
\node[gnet](G) [right = 1cm of I, label=center: Gnet]{};
\node[output2](FIM)[right = 1cm of G, label=center: Fake images]{};
\node[dnet](D)[below right = 0.1cm and 1cm of FIM, label=center:Dnet]{};
\node[dnet](O) [right = 1cm of D, label=center: $\mathbf{z}^{(\mathtt{O})}$]{};
\node[](fake)[above right  = 0.1cm and 1cm of O)]{\raisebox{-.3cm}{fake}};
\node[](RIM) [below = 1cm of FIM, label=center: Real images]{};
\node[](real)[below = 0.8cm of fake]{\raisebox{-.3cm}{real}};
\path (I) edge[Forward, thin] (G)
	(G) edge[Forward, thin] (FIM)
	(FIM) edge[Forward, thin, out=0, in = 180] (D)
	(D) edge[Forward, thin, out=0, in=180] (O)
	(O) edge[Forward, thin, out=0, in =180] ([xshift=0mm]fake.west)
	([xshift=10mm]RIM.east) edge[Forward, thin, out=0, in=180] (D)
	(O) edge[Forward, thin, out=0, in=180] (real) 
	(real) edge[Inference, thin, bend left= 40] (O)
	  (fake) edge[Inference, thin, bend right= 40] (O)
	 (O) edge[Inference, thin, bend left= 40] (D);
\end{tikzpicture}}}
\\\bigskip
\subfloat[Generator]{\scalebox{0.9}{\begin{tikzpicture}
\node[input](I)[label=center: Noise]{};
\node[gnet](G) [right = 1cm of I, label=center: Gnet]{};
\node[output2](FIM)[right = 1cm of G, label=center: Fake images]{};
\node[dnet](D)[right = 1cm of FIM, label=center:Dnet]{};
\node[dnet](O) [right = 1cm of D, label=center: $\mathbf{z}^{(\mathtt{O})}$]{};
\node[](real)[right = 1cm of O]{\raisebox{-.1cm}{real}};
\path	 (I) edge[Forward, thin] (G)
	(G) edge[Forward, thin] (FIM)
	(FIM) edge[Forward, thin] (D)
	(D) edge[Forward, thin] (O)
	(O) edge[Forward, thin] (real)
	(real) edge[Inference, thin, bend left= 60] (O)
	(O) edge[Inference, thin, bend left= 60] (D)
	(D) edge[Inference, thin, bend left= 50] (FIM)
	 (FIM) edge[Inference, thin, bend left= 50] (G);
\end{tikzpicture}}}
\caption{Schematic representation for (a) the discriminator and (b) the generator network of a GAN. The magenta arrows correspond to the flow of information during the forward propagation of uncertainty through the network and the blue arrows represent the flow of information during the inference procedure.}
\label{fig:gan}
\end{center}
\end{figure}
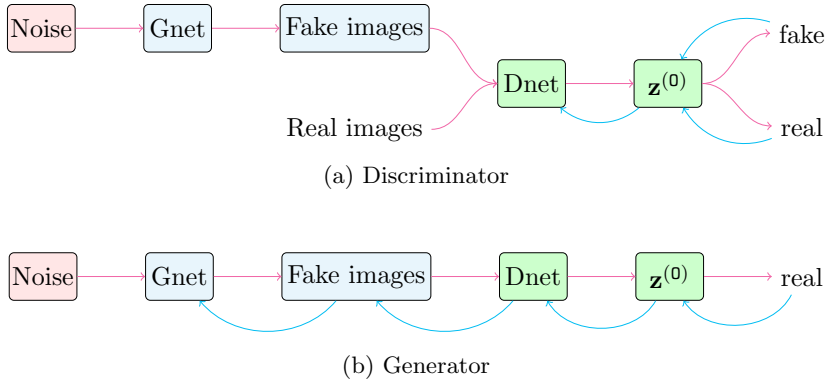

GANs build on a generative approach to create realizations from the input covariates distribution $f(\bm{x})$. In practice, the input $\bm{x}$ typically takes the form of images. The GAN architecture consists in a discriminator (Dnet) and a generator (Gnet) network as depicted in Figure \ref{fig:gan}. 
The generator network is used to create synthetic data and the discriminator is a binary classification network that allows distinguishing synthetic (i.e. fake) and real data. The generator takes random noise as input. 
Figure \ref{fig:gan} presents the schematic representation of a GAN where the magenta arrows depict the dependencies in the network and correspond to the flow of information during the forward propagation of uncertainty through the network. The blue arrows running backward represent the flow of information during the inference procedure. In Figure \ref{fig:gan}a, the Dnet takes as input either a fake image generated from an input noise sample that is passed to the Gnet, or a real image; once passed through the Dnet, that image is either classified as fake or real in the output layer $\bm{z}^{(\mathtt{O})}$. The parameters of the Dnet are then updated in the inference step by relying on the true label for that image. During this inference step, the parameters of the Gnet are not updated. For the generator presented in Figure \ref{fig:gan}b, an input noise sample is passed to the Gnet in order to return a fake image that is then passed through the Dnet. We then update the Gnet's parameters by forcing the observation for the output layer of the Dnet, $\bm{z}^{(\mathtt{O})}$, to be classified as a real image. Note that although the inference procedure passes through the Dnet, only the parameters from the Gnet are updated for the generator.

Information maximizing generative adversarial networks (InfoGAN) \cite{chen2016infogan} allow controlling the latent space by disentangling interpretable data representations such as classes, rotation, and width. As a GAN, an infoGAN consists in two networks: discriminator and generator. With infoGAN, the discriminator network has two purposes: (1) distinguishing the fake and real data, and (2) learn the interpretable  representations. Figure \ref{fig:infoGAN}a shows the architecture of the discriminator where the Pnet is employed to classify the real and fake data just like in normal GANs, while the Qnet allows predicting the latent variables associated with an input image.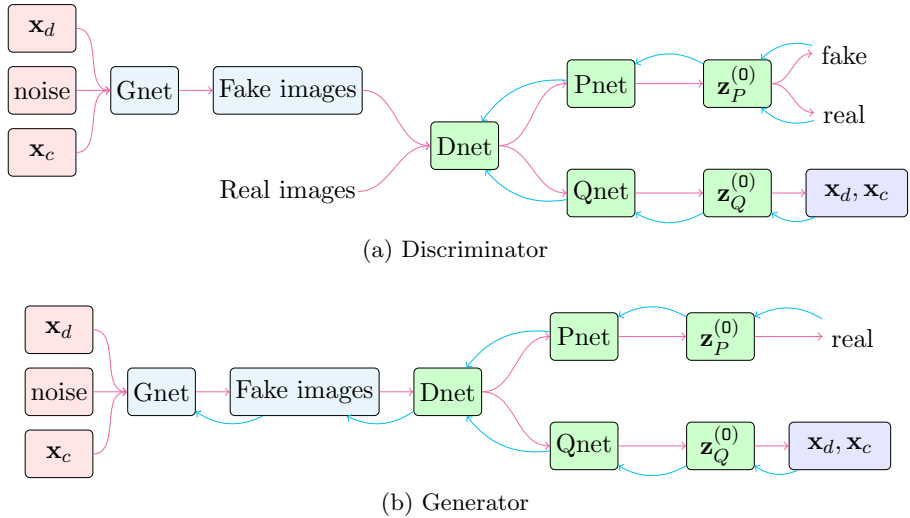
\begin{figure}[t]
\begin{center}
\input{tikz_setup_GAN.tex}
\subfloat[Discriminator]{\scalebox{0.9}{\begin{tikzpicture}
\node[input](I)[label=center: noise]{};
\node[input](I2)[below=0.2cm of I, label=center:$\mathbf{x}_{c}$]{};
\node[input](I3)[above=0.2cm of I, label=center:$\mathbf{x}_{d}$]{};
\node[gnet](G) [right = 0.5cm of I, label=center: Gnet]{};
\node[output2](FIM)[right = 0.5cm of G, label=center: Fake images]{};
\node[dnet](D)[below right = 0.1cm and 1cm of FIM, label=center:Dnet]{};
\node[dnet](P) [above right = 0.2cm and 1 cm of D, label=center: Pnet]{}; 
\node[dnet](OP) [right = 1cm of P, label=center: $\mathbf{z}^{(\mathtt{O})}_{P}$]{};
\node[](fake)[above right  = -0.2cm and 0.6cm of OP]{\raisebox{-.3cm}{fake}};
\node[dnet](Q) [below= 0.9cm of P, label=center: Qnet]{}; 
\node[dnet](OQ) [right = 1cm of Q, label=center: $\mathbf{z}^{(\mathtt{O})}_{Q}$]{};
\node[obs](O) [right= 0.5cm of OQ]{$\mathbf{x}_{d}, \mathbf{x}_{c}$}; 
\node[](RIM) [below = 1cm of FIM, label=center: Real images]{};
\node[](real)[below = 0.3cm of fake]{\raisebox{-.3cm}{real}};

\path (I) edge[Forward, thin] (G)
	(I2) edge[Forward, thin, out = 0, in = 180] (G)
	(I3) edge[Forward, thin, out = 0, in = 180] (G)
	(G) edge[Forward, thin] (FIM)
	(FIM) edge[Forward, thin,  out=0, in=180] (D)
	(D) edge[Forward, thin, out=0, in=180] (P)
	(P) edge[Forward, thin, out=0, in=180] (OP)
	(OP) edge[Forward, thin, out=0, in=180] (fake)
	(D) edge[Forward, thin, out=0, in=180] (Q)
	(Q) edge[Forward, thin, out=0, in=180] (OQ)
	(OQ) edge[Forward, thin, out=0, in=180] (O)
	([xshift=9mm]RIM.east) edge[Forward, thin, out=0, in=180] (D)	
	(OP) edge[Forward, thin, out=0, in=180] (real)
	(real) edge[Inference, thin, bend left= 30] (OP)
	 (fake) edge[Inference, thin, bend right= 30] (OP)
	 (OP) edge[Inference, thin, bend right= 30] (P)
	 (P) edge[Inference, thin, bend right= 30] (D)
	(O) edge[Inference, thin, bend left= 30] (OQ)
	(OQ) edge[Inference, thin, bend left= 30] (Q)
	(Q) edge[Inference, thin, bend left= 30] (D);

\end{tikzpicture}}}
\\\smallskip
\subfloat[Generator]{\scalebox{0.9}{\begin{tikzpicture}
\node[input](I)[label=center: noise]{};
\node[input](I2)[below=0.2cm of I, label=center:$\mathbf{x}_{c}$]{};
\node[input](I3)[above=0.2cm of I, label=center:$\mathbf{x}_{d}$]{};
\node[gnet](G) [right = 0.5cm of I, label=center: Gnet]{};
\node[output2](FIM)[right = 0.5cm of G, label=center: Fake images]{};
\node[dnet](D)[right = 0.5cm of FIM, label=center:Dnet]{};
\node[dnet](P) [above right = 0.1cm and 1 cm of D, label=center: Pnet]{}; 
\node[dnet](OP) [right = 1cm of P, label=center: $\mathbf{z}^{(\mathtt{O})}_{P}$]{};
\node[](real)[right = 1cm of OP]{\raisebox{-.3cm}{real}};
\node[dnet](Q) [below= 0.9cm of P, label=center: Qnet]{}; 
\node[dnet](OQ) [right = 1cm of Q, label=center: $\mathbf{z}^{(\mathtt{O})}_{Q}$]{};
\node[obs](O) [right= 0.5cm of OQ]{$\mathbf{x}_{d}, \mathbf{x}_{c}$}; 
\path (I) edge[Forward, thin] (G)
	(I2) edge[Forward, thin, out = 0, in = 180] (G)
	(I3) edge[Forward, thin, out = 0, in = 180] (G)
	(G) edge[Forward, thin] (FIM)
	(FIM) edge[Forward, thin] (D)
	(D) edge[Forward, thin, out=0, in=180] (P)
	(P) edge[Forward, thin, out=0, in=180] (OP)
	(OP) edge[Forward, thin, out=0, in=180] (real)
	(D) edge[Forward, thin, out=0, in=180] (Q)	
	(Q) edge[Forward, thin, out=0, in=180] (OQ)
	(OQ) edge[Forward, thin, out=0, in=180] (O)
        (real) edge[Inference, thin, bend right= 30] (OP)
	(OP) edge[Inference, thin, bend right= 30] (P)
	(P) edge[Inference, thin, bend right= 30] (D)
	(O) edge[Inference, thin, bend left= 30] (OQ)
	(OQ) edge[Inference, thin, bend left= 30] (Q)
	(Q) edge[Inference, thin, bend left= 30] (D)
	(D) edge[Inference, thin, bend left = 30] (FIM)
	(FIM) edge[Inference, thin, bend left= 30] (G);
\end{tikzpicture}}}
\caption{Schematic representation for (a) the discriminator and (b) the generator network of an infoGAN. The magenta arrows correspond to the flow of information during the forward propagation of uncertainty through the network and the blue arrows represent the flow of information during the inference procedure.}
\label{fig:infoGAN}
\end{center}
\end{figure} Both the Pnet and Qnet are outputs from the shared Dnet taking as input either a real or fake image that has been created by the Gnet. During inference, the parameters of the Dnet, Pnet and Qnet are updated by relying on the true label for that image {as well as the latent variables in the case of fake images. The inputs of the generator networks consist in an input noise sample, a categorical variable vector ($\mathbf{x}_{d}$), and continuous variable vector ($\mathbf{x}_{c}$). For example, in the case of the MNIST dataset \cite{lecun1998gradient}, the discrete variables represent the digits from 0 to 9 and the continuous variable can be the width and rotation of each digit. In practice, $\mathbf{x}_{d}$ and $\mathbf{x}_{c}$ are randomly generated during training and can be specifically queried once the training is completed. Note that again only the Gnet's parameters are updated in Figure \ref{fig:infoGAN}b, even if during inference, the information has to flow backward through the Pnet, Qnet and Dnet.

For image datasets, GANs architectures commonly employ both CNN and FNN which involve the operations such as multiplication, addition and non-linear transformation. Therefore, TAGI's mathematical developments can be directly applied to perform the operations in GAN and infoGAN architectures without having to rely on any gradient whatsoever, or to develop any new equations other than those presented in Section \ref{SS:CNN} for CNNs.

%

\section{Benchmarks}\label{S:Benchmarks} 
In this section, we compare the performance of TAGI for deep neural networks with existing deterministic or Bayesian neural networks trained with backpropagation. In the case of TAGI, the hyperparameters are kept the same for all benchmarks with a batch size $\mathtt{B}=16$, a maximal number of epochs $\mathtt{E}=50$, a noise discount factor $\eta=0.975$, and all weights are initialized using the He's approach \cite{he2015delving}. These hyperparameters are kept constant across all benchmarks to demonstrate the robustness of the approach towards them. The only hyperparameter specific to each experiment is the initial observation noise standard deviation $\sigma_{V}^{0}$. Additionally, for the MNIST  \cite{lecun1998gradient}  and CIFAR10 \cite{krizhevsky2009learning} datasets, the input data was first normalized between 0 and 1 and then the mean of the normalized data was subtracted. For the celebA dataset \cite{liu2015faceattributes}, we preprocess the input data by subtracting the mean and dividing by the standard deviation of the input data.
\subsection{MNIST}
The first benchmark is the MNIST digit classification task. Table \ref{TAB_MNIST} compares the classification accuracy with the results from Wan et al. \cite{wan2013regularization} where both approaches use the same CNN architecture with 2 convolutional layers (32-64) and a fully connected layer with 150 hidden units. For this experiment, the initial observation noise standard deviation $\sigma_{V}^{0}=1$. Results show that TAGI can already reach a classification error $<2\%$ after the first epoch and after $\mathtt{E}=50$ epochs, the performance is slightly better than the results obtained with backpropagation trained with over 20 times as many epochs. 
\begin{table}[htbp]
\centering
\caption{Comparison of the classification performance on MNIST between a network trained with TAGI and backpropagation (BP). ($\mathtt{E}$, number of epochs, $\mathtt{B}$, batch size)}
\begin{tabular}{ccc|cc}
    	\addlinespace
    \toprule
    	\multicolumn{1}{l}{}&\multicolumn{2}{c}{Error Rate [\%]}&\multicolumn{2}{|c}{Hyperparameters}\\[0pt] 
	\cmidrule{2-5}
$\mathtt{}$&$e=1$& $e=\mathtt{E}$ &$\mathtt{E}$&$\mathtt{B}$\\[0pt]   
\cmidrule{1-5}
\multirow{1}{*}{\bf TAGI}&$1.88$ &$0.64$ &$50$ &$16$\\[2pt]
BP~\cite{wan2013regularization}&- &$0.67$ &$1000$ &$128$\\[2pt]
\bottomrule
\end{tabular}
\label{TAB_MNIST}
\end{table}

In a second experiment on the same dataset, we employ an infoGAN (see \S \ref{ss:gan}) in order to test the performance of TAGI on unsupervised learning tasks. The hyperparameters specific to this experiment are the two observation noise initial standard deviations for the output of the Pnet and Qnet that are taken as $\sigma_{V}^{0, P} = \sigma_{V}^{0, Q} = 3$. As a reference, we employ the input latent space and infoGAN architecture described in \cite{chen2016infogan}. The input latent space includes $62$ Gaussian noise variables, one categorical variable ${x}_{d}$ containing ten classes, and two continuous variables for representing the width and rotation of the digit. In the case of TAGI, we employ the same input latent space with a modification to the architecture. Specifically, for the Gnet, we use half as many convolutional filters and half as many hidden units in the output's fully connected layer. For the Qnet, we use a fully connected layer containing 300 hidden units instead of 128. Overall, our architecture contains $\approx$ 2.1M parameters versus 13.3M for the network used as reference \cite{chen2016infogan}, and we train over 50 epochs whereas they train with 100 epochs. The complete architecture's details are provided in Appendix \ref{BG:mnist}.

The results obtained using TAGI are presented in Figures \ref{FIG:IG_minst}a--c while the results obtained with backpropagation are shown in Figures \ref{FIG:IG_minst}d--f. Note that the results for the MNIST trained with backpropagation were obtained using the code from \cite{InfoGAN-PyTorch}. Overall, we denote an advantage for the network trained with backpropagation as TAGI can only disentangle 9 out of 10 digits. Moreover, for TAGI, the two continuous variables forming the latent space do not have an effect as clear as in the case of backpropagation. The reason for this sub-par performance is that for TAGI, the output layer uncertainty is homoscedastic whereas, in the case of \cite{chen2016infogan}, it is heteroscesastic so that they optimize both the expected values and variances for the two continuous variables on the output layer. With its current formulation TAGI can only handle homoscedastic output uncertainty which jeopardizes its capacity at disentangling the latent space with the unsupervised setup where we learn both the categorical and continuous variables at once. 
\begin{figure}[htbp]\centering
\subfloat[Reference -- TAGI]{\includegraphics[width=38mm]{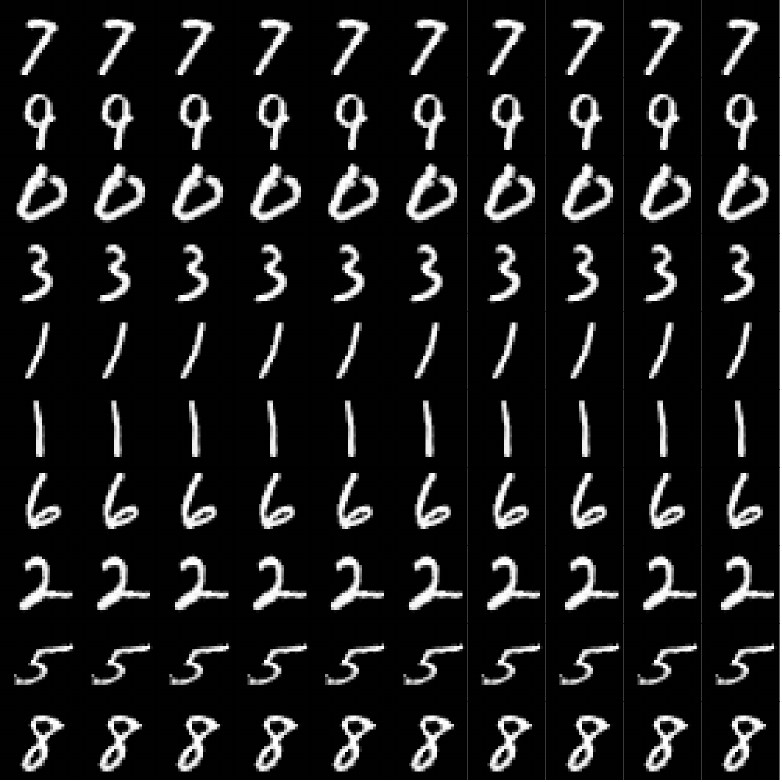}}
\hspace{0.5cm}\subfloat[Width -- TAGI]{\includegraphics[width=38mm]{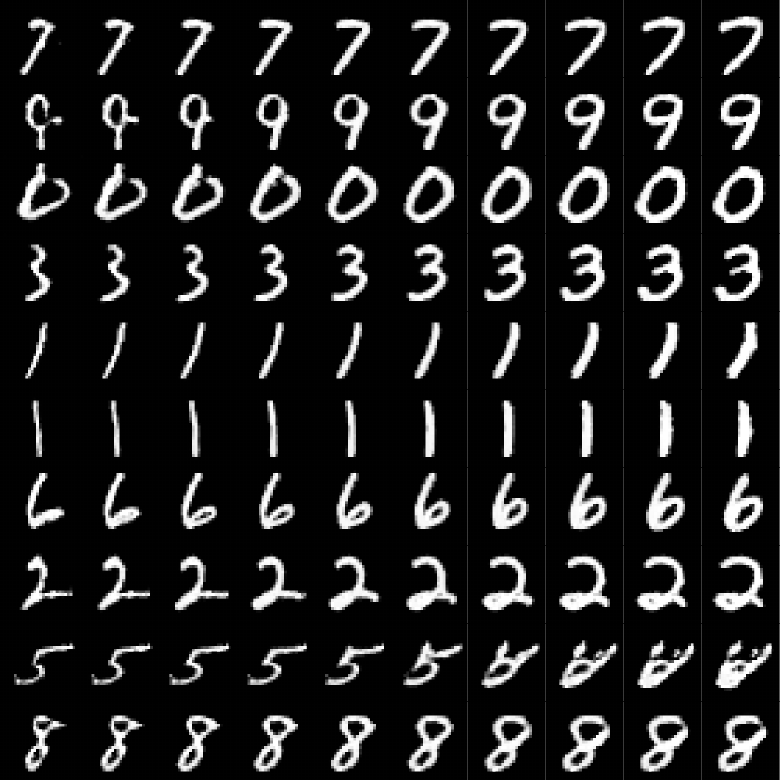}}\hspace{0.5cm}\subfloat[Rotation -- TAGI]{\includegraphics[width=38mm]{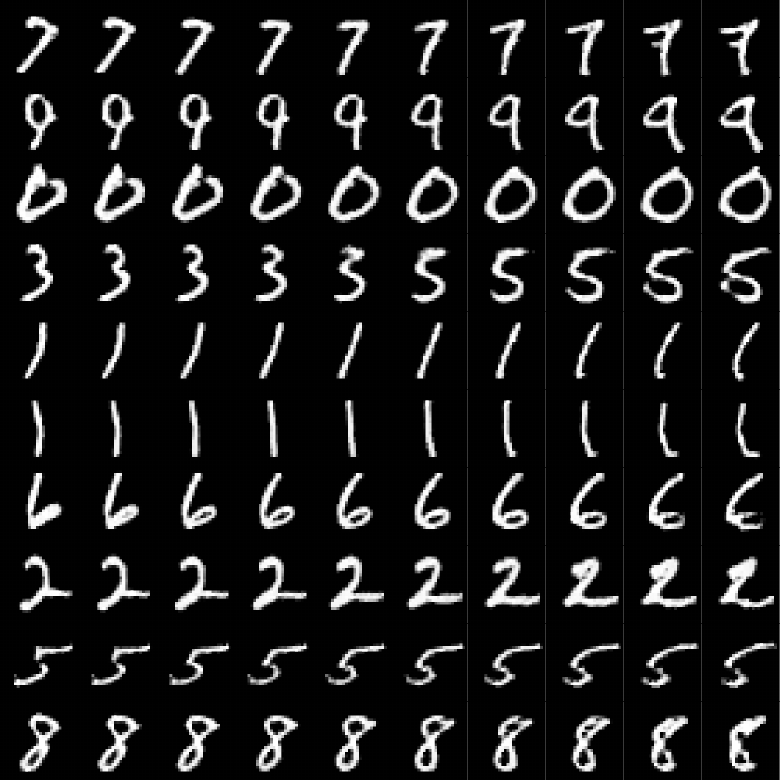}}\\[8pt]
\subfloat[Reference -- Backprop]{\includegraphics[width=40mm]{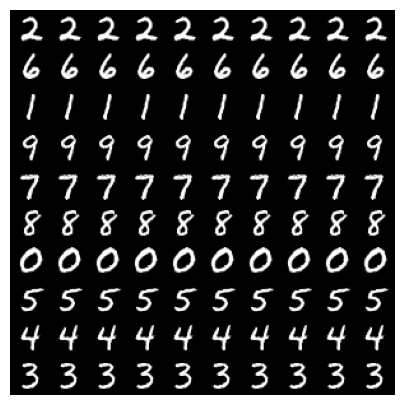}}
\hspace{0.3cm}\subfloat[Width -- Backprop]{\includegraphics[width=40mm]{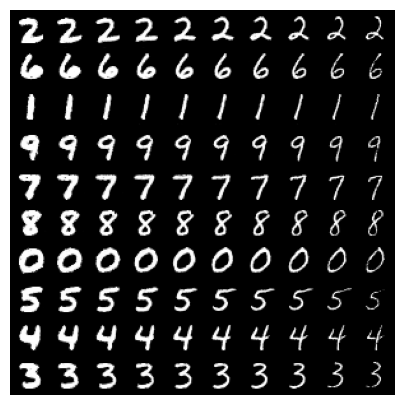}}\hspace{0.3cm}\subfloat[Rotation -- Backprop]{\includegraphics[width=40mm]{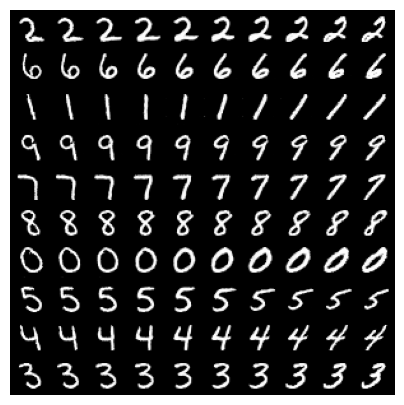}}
\caption{Comparison of the performance of a backpropagation and TAGI trained InfoGAN applied to the MNIST dataset. Results present reference digit generation (a,d), and their control with a latent space consisting of the width (b,e) and rotation (c,f).}
\label{FIG:IG_minst}
\end{figure}

\subsection{CIFAR-10}
The second benchmark is the CIFAR-10 image classification task. Experiments are conducted for a simple network with three convolutional layers, as well as for the Resnet18 architecture \cite{he2016deep}. For both cases, we also experimented with the usage of layer and batch normalization. The details regarding the 3-layers convolutional network are presented in Appendix \ref{C:cifar10)}. Note that we only apply the data augmentation techniques including random cropping and horizontal flipping \cite{osawa2019practical} to the Resnet18 architecture for the comparison purpose with the neural networks trained with backpropagation. In the case of TAGI, the initial observation noise standard deviation was set to the same value as for the MNIST experiment, i.e., $\sigma_{V}^{0}=1$. Table \ref{TAB:CIFAR10} presents the classification error rate for TAGI-trained networks as well as the results from several papers using deterministic as well as Bayesian methods for the same network architectures. Note that in the case of Resnet18 network, the results presented are the average from three runs. 

\begin{table}[h!]
\centering
\caption{Comparison of the classification performance on CIFAR-10 between a network trained with TAGI and backpropagation (BP), BP with layer normalization (-LN), BP with batch normalization (-BN), MC-dropout, and VOGN. ($\mathtt{E}$: number of epochs; $\mathtt{B}$: batch size ; DA: data augmentation).}
\begin{tabular}{cccc|cc}    	\addlinespace
    \toprule
    	\multicolumn{1}{l}{}&\multicolumn{1}{l}{}&\multicolumn{2}{c}{Error rate [\%]}&\multicolumn{2}{|c}{Hyperparameters}\\[0pt] 
	\cmidrule{2-6}
Architecture&Method&$e=1$& $e=\mathtt{E}$ &$\mathtt{E}$&$\mathtt{B}$\\[0pt]   
\cmidrule{1-6}
\multirow{7}{*}{3 conv.}&\bf TAGI&$44.3$ &$23.6$ &$50$ &16\\[2pt]
&{\bf TAGI}-LN&41.4 &$22.6$ &$50$ &$16$\\[2pt]
&{\bf TAGI}-BN&42.6 &$22.0$ &$50$ &$16$\\[2pt]
&BP~\cite{wan2013regularization}&- &$23.5$ &$150$ &$128$\\[2pt]
&BP-LN~\cite{ren2016normalizing}&- &$27.9$ &$100$ &$100$\\[2pt]
&BP-BN~\cite{ren2016normalizing}&- &$21.9$ &$100$ &$100$\\[2pt]
\cmidrule{1-6}
\multirow{4.5}{*}{Resnet18}&\bf TAGI&$65.9$ &$13.8$ &$50$ &16\\[2pt]
\multirow{4.5}{*}{(DA)}&BP~\cite{osawa2019practical}&- &$14.0$ &$160$ &$128$\\[2pt]
&MC-dropout~\cite{osawa2019practical}&- &$17.2$ &$161$ &$128$\\[2pt]
&VOGN~\cite{osawa2019practical}&- &$15.7$ &$161$ &$128$\\[2pt]
\cmidrule{1-6}
\multirow{1}{*}{\begin{tabular}{c}Resnet18half\\[2pt](DA)\end{tabular}}&& &&&\\[-4pt]
\multirow{-1}{*}{}&\bf TAGI&$65.9$ &$14.9$ &$50$ &16\\[6pt]
\bottomrule
\end{tabular}
\label{TAB:CIFAR10}
\end{table}
First, for the 3-layer convolutional architecture, the results show that TAGI matches the SOTA performance while using a lower number of epochs for the experiments with the batch normalization and without normalization. TAGI outperforms the SOTA performance for the experiment with the layer normalization.  Second, for the Resnet18 architecture, TAGI outperforms deterministic and Bayesian approach with a third of the epochs for training. Note that the performance of TAGI still matches the performance of MC-dropout and VOGN even if we employ half as many convolutional filters in the Resnet18 architecture, i.e., Resnet18half.

Table \ref{TAB:CIFAR10_calibration} presents the predictive uncertainty calibration metrics \cite{10.5555/3305381.3305518} for the Resnet18 and Resnet18half architecture. The results show that for the Resnet18, TAGI performs poorly on the NLL and ECE criteria, yet, for the same criteria, it performs on par with other methods for the Resnet18half. One of reason behind this limitation is that TAGI is currently limited to using homoscedastic observation noise parameters $\sigma_V$, which is not the case for existing methods relying on backpropagation.  Despite this limitation, none of the backpropagation-based methods match the AUROC criterion obtained with TAGI.
\begin{table}[h!]
\centering
\caption{Comparison of the predictive uncertainty calibration performance on CIFAR-10 between a network trained with TAGI and backpropagation (BP), BP, MC-dropout, and VOGN. ($\mathtt{E}$: number of epochs; $\mathtt{B}$: batch size ; NLL: Negative Log-Likelihood; ECE: Expected Calibration Error; AUROC: Area Under the Receiver Operating Characteristics). See Table \ref{TAB:CIFAR10_std_cali} for standard deviations.}
\scalebox{0.9}{\begin{tabular}{ccccc|cc}    	\addlinespace
    \toprule
    	\multicolumn{2}{l}{}&\multicolumn{1}{c}{NLL}&\multicolumn{1}{c}{ECE}&\multicolumn{1}{c}{AUROC}&\multicolumn{2}{|c}{Hyperparameters}\\[0pt] 
	\cmidrule{2-7}
Architecture&Method&$e=\mathtt{E}$&$e=\mathtt{E}$ &$e=\mathtt{E}$&$\mathtt{E}$&$\mathtt{B}$\\[0pt]   
\cmidrule{1-7}
\multirow{4.5}{*}{Resnet18}&\bf TAGI &0.588& 0.143&0.980&$50$ &16\\[2pt]
\multirow{4.5}{*}{(DA)}&BP~\cite{osawa2019practical}&0.55&0.082& 0.877&$160$ &$128$\\[2pt]
&MC-dropout~\cite{osawa2019practical} &0.51&0.166& 0.768  &$161$ &$128$\\[2pt]
&VOGN~\cite{osawa2019practical} &0.477&0.04& 0.876  &$161$ &$128$\\[2pt]
\cmidrule{1-7}
\multirow{1}{*}{\begin{tabular}{c}Resnet18half\\[2pt](DA)\end{tabular}}&\bf && &&$$ &\\[-4pt]
&\bf TAGI &0.541& 0.046&0.978&$50$ &16\\[6pt]
\bottomrule
\end{tabular}}
\label{TAB:CIFAR10_calibration}
\end{table}

\subsection{CelebA}
In the last experiment, we applied infoGAN to the CelebA dataset for which we downsampled the original images from $3\times224\times224$ to $3\times32\times32$. The hyperparameters specific to this experiment are the two observation noise initial standard deviations for the output of the Pnet and Qnet that are taken as $\sigma_{V}^{0, P} = 3$ and $\sigma_{V}^{0,Q} = 8$.  As a reference, we use the input latent space and the infoGAN architecture described in \cite{chen2016infogan}. The input latent space contains a Gaussian noise vector including $128$ variables along with ten categorical variables $\bf{x}_{d}$ each having ten classes. The specifications of the architecture used with TAGI is presented in Appendix \ref{BG:celebA}. Overall the TAGI-trained architecture contains approximately 0.7M parameters in comparison with 4.1M for the architecture which serves here as reference. Results for the infoGan trained with backpropagation were obtained using the code from \cite{InfoGAN-PyTorch}.

Figure \ref{FIG:celebA_1} compares a sample of the training images with those generated  with both the architecture from \cite{chen2016infogan} trained with backpropagation over 100 epochs, and with an architecture containing six times fewer parameters trained with TAGI over 50 epochs. These results show that despite using a smaller network, TAGI performs at par with the larger network trained with backpropagation. 
\begin{figure}[h!]\centering
\subfloat[Sample training images from the CelebA dataset.]{\includegraphics[width=40mm]{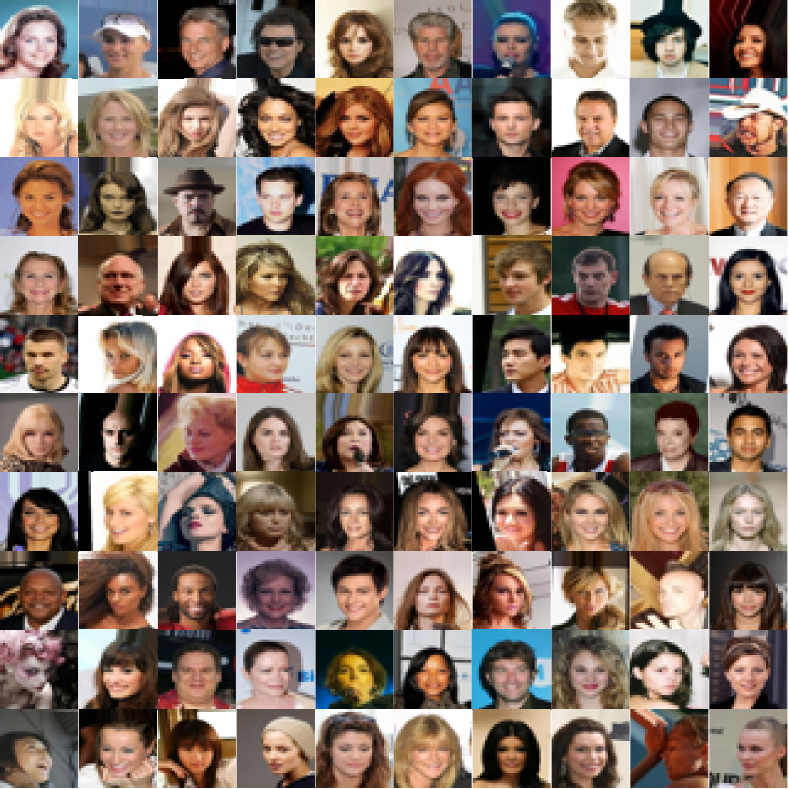}}\hspace{0.25cm}
\subfloat[Backpropagation (100 epochs)]{\includegraphics[width=40mm]{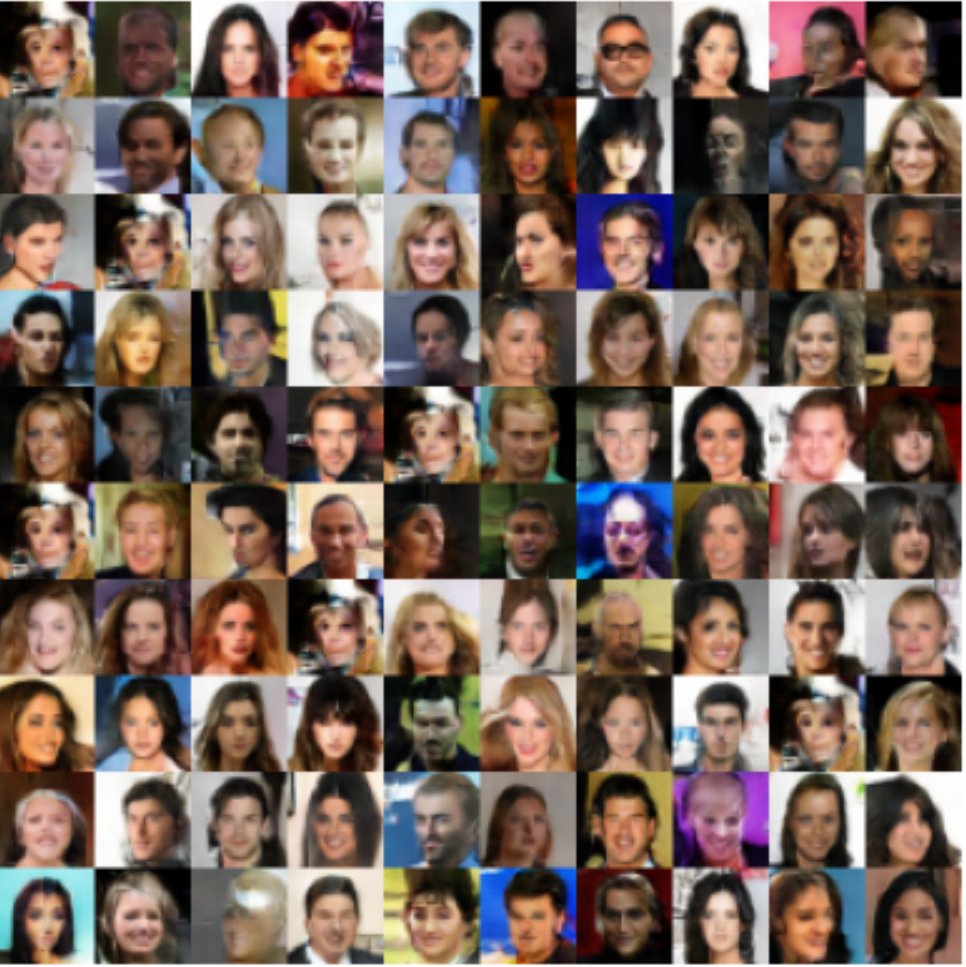}}
\hspace{0.25cm}\subfloat[TAGI (50 epochs)]{\includegraphics[width=40mm]{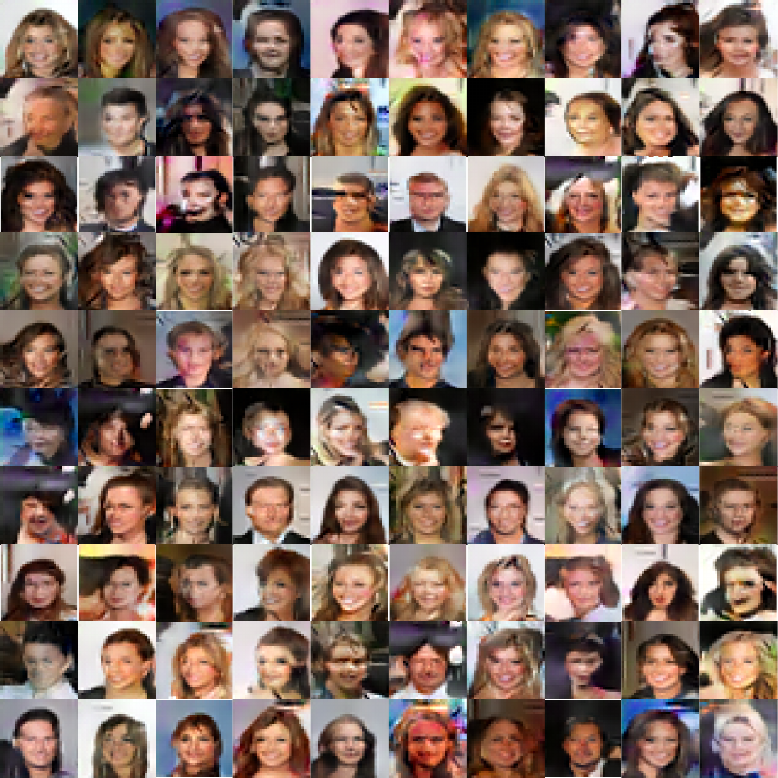}}
\caption{Comparison of a sample of training images (a) with the synthetic images generated using infoGAN with (b) the architecture presented in \cite{chen2016infogan} trained with backpropagation, and with (c) a network six times smaller trained with TAGI.}
\label{FIG:celebA_1}
\end{figure}

Figure \ref{FIG:celebA_2} presents the variations in images for the latent variable controlling the hair colour. Again, TAGI performs at par with the larger network trained with backpropagation. Other latent variables such as the gender, azimuth, hairstyle, and contrasts are presented in Appendix \ref{Appendix:C}

\begin{figure}[h!]\centering
\subfloat[Backpropagation (100 epochs)]{\includegraphics[width=60mm]{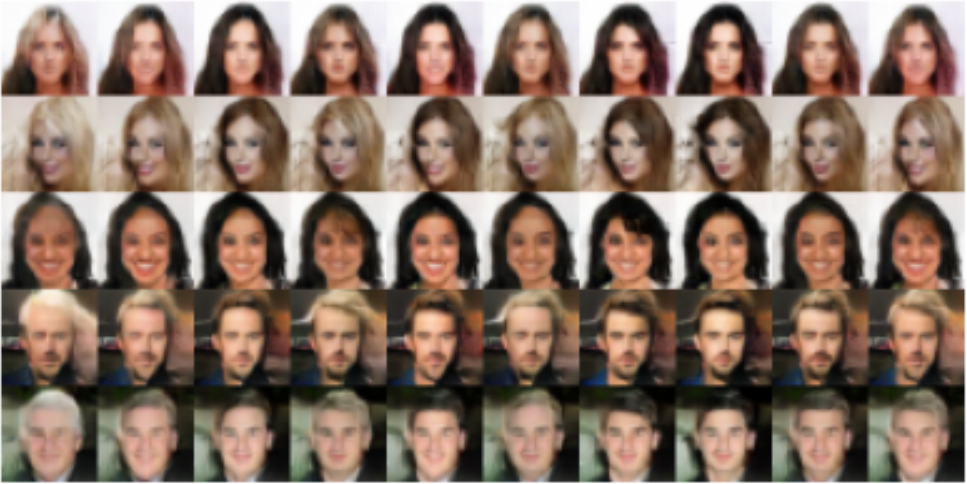}}
\hspace{0.3cm}\subfloat[TAGI (50 epochs)]{\includegraphics[width=60mm]{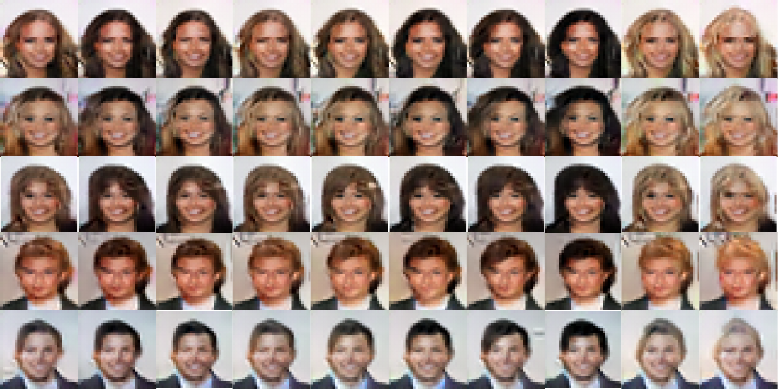}}
\caption{Comparison of the capacity to control the latent space for hair colour for (a) the architecture presented in \cite{chen2016infogan} and trained with backpropagation, and (b) a network six times smaller trained with TAGI.}
\label{FIG:celebA_2}
\end{figure}


\section{Discussion}
For the infoGAN experiments, the results have shown that TAGI can reach a performance at par with a network train with backpropagation while using significantly smaller architectures and a lower number of epochs. We opted for this approach because TAGI's implementation is not yet optimized for computational efficiency so the current computing time is approximately one order of magnitude higher than when training the same deterministic neural network using backpropagation.

Throughout the experiments, TAGI employs far fewer hyperparameters than traditional neural networks trained with backpropagation. Moreover, TAGI's robustness toward these hyperparameter values allows keeping most of them constant across various tasks. The main hyperparameters that still need to be adapted with the size of the network employed are the initial output noise standard deviations, $\sigma_V^0$. In the future, these hyperparameters could be passed as hidden states that are learned online like other weights and bias, while simultaneously adding the capacity to model heteroscedastic output noise variance.   

In addition, we can improve the classification performance with the same architecture by using smaller batch sizes, which, however, decreases the potential for parallelization. The same applies, for larger batch sizes where the losses in classification performance is compensated by an increase in computation parallelization. A systematic study of the effect of batch sizes remains to be explored.

\section{Conclusion}
This paper demonstrates the applicability of TAGI to deep neural networks. The results obtained from different tasks confirm that TAGI is able to perform analytical inference for the parameters of deep neural network. Although the computational efficiency is still below that of standard approaches relying on backpropagation, TAGI outperforms these approaches on classification tasks and match the performance for infoGANs while using smaller architectures trained with fewer epochs. Future code developments oriented toward efficiency should reduce that gap.

\subsubsection*{Acknowledgments}
The first author was financially supported by research grants from Hydro-Quebec, and the Natural Sciences and Engineering Research Council of Canada (NSERC).

\bibliographystyle{plain}
\small
\bibliography{Goulet_reference_librairy}

\newpage
\section*{Appendix}
\appendix
\section{Cross-covariances for layer normalization}\label{Appendix:A}
Assuming that $\bm{Z}$, $\bm{W}$, and $\bm{B}$ are independent on each other,
\begin{equation}
\begin{array}{rcl}
\text{cov}\left(\bm{Z}, \bm{Z}^{+}\right)&=&\text{cov}\left(\bm{Z}, \tfrac{1}{{\sigma}_{\bm{A}}}\bm{W}\bm{A}- \tfrac{{\mu}_{\bm{A}}}{{\sigma}_{\bm{A}}}\bm{W} + \bm{B}\right)\\[10pt]

&=&\tfrac{1}{{\sigma}_{\bm{A}}}\text{cov}\left(\bm{Z}, \bm{W}\bm{A}\right) - \cancelto{0}{\tfrac{{\mu}_{\bm{A}}}{{\sigma}_{\bm{A}}}\text{cov}\left(\bm{Z}, \bm{W}\right)} + \cancelto{0}{\text{cov}\left(\bm{Z}, \bm{B}\right)}\\[16pt]

&=&\tfrac{1}{{\sigma}_{\bm{A}}}\left[\text{cov}\left(\bm{Z}, \bm{A}\right)\bm\mu_{\bm{W}} + \cancelto{0}{\text{cov}\left(\bm{Z}, \bm{W}\right)\bm\mu_{\bm{A}}}\right]\\[16pt]

&=&\tfrac{1}{{\sigma}_{\bm{A}}}\text{cov}\left(\bm{Z}, \bm{Z}\right)\mathbf{J}\bm\mu_{\bm{W}}, 
\end{array}
\end{equation}
where $\bm{A}=\mathbf{J}\left(\bm{Z}-\bm\mu_{\bm{Z}}\right) + \sigma(\bm{Z})$ with $\sigma(.)$ being the activation function.

\begin{equation}
\begin{array}{rcl}
\text{cov}\left(\bm{W}, \bm{Z}^{+}\right)&=&\text{cov}\left(\bm{W}, \tfrac{1}{{\sigma}_{\bm{A}}}\bm{W}\bm{A}- \tfrac{{\mu}_{\bm{A}}}{{\sigma}_{\bm{A}}}\bm{W} + \bm{B}\right)\\[10pt]

&=&\tfrac{1}{{\sigma}_{\bm{A}}}\text{cov}\left(\bm{W}, \bm{W}\bm{A}\right) - \tfrac{{\mu}_{\bm{A}}}{{\sigma}_{\bm{A}}}\text{cov}\left(\bm{W}, \bm{W}\right) + \cancelto{0}{\text{cov}\left(\bm{W}, \bm{B}\right)}\\[16pt]

&=&\tfrac{1}{{\sigma}_{\bm{A}}}\left[\cancelto{0}{\text{cov}\left(\bm{W}, \bm{A}\right)\bm\mu_{\bm{W}}} + \text{cov}\left(\bm{W}, \bm{W}\right)\bm\mu_{\bm{A}}\right] - \tfrac{{\mu}_{\bm{A}}}{{\sigma}_{\bm{A}}}\text{cov}\left(\bm{W}, \bm{W}\right)\\[16pt]

&=&\tfrac{1}{{\sigma}_{\bm{A}}}\text{cov}\left(\bm{W}, \bm{W}\right)\bm\mu_{\bm{A}} - \tfrac{{\mu}_{\bm{A}}}{{\sigma}_{\bm{A}}}\text{cov}\left(\bm{W}, \bm{W}\right). 
\end{array}
\end{equation}

\begin{equation}
\begin{array}{rcl}
\text{cov}\left(\bm{Z}, \bm{Z}^{+}\right)&=&\text{cov}\left(\bm{B}, \tfrac{1}{{\sigma}_{\bm{A}}}\bm{W}\bm{A}- \tfrac{{\mu}_{\bm{A}}}{{\sigma}_{\bm{A}}}\bm{W} + \bm{B}\right)\\[10pt]

&=&\cancelto{0}{\tfrac{1}{{\sigma}_{\bm{A}}}\text{cov}\left(\bm{B}, \bm{W}\bm{A}\right)} - \cancelto{0}{\tfrac{{\mu}_{\bm{A}}}{{\sigma}_{\bm{A}}}\text{cov}\left(\bm{B}, \bm{W}\right)} + \text{cov}\left(\bm{B}, \bm{B}\right)\\[16pt]

&=&\text{cov}\left(\bm{B}, \bm{B}\right).
\end{array}
\end{equation}

\section{Model Architectures}\label{Appendix:B}
This appendix contains the specifications for each model architecture in the experiment section. $D$ refers to a layer depth; $W$ refers to a layer width; $H$ refers to the layer height in case of convolutional or pooling layers; $K$ refers to the kernel size; $P$ refers to the convolutional kernel padding; $S$ refers to the convolution stride; $\sigma$ refers to the activation function type; ReLU refers to rectified linear unit; lReLU refers to leaky rectified linear unit.
\subsection{Classification}\label{B:classificaiton}
\subsubsection{MNIST}\label{C:mnist}

\begin{table}[h!]
\begin{center}
\caption{Configuration details for the CNN applied to the MNIST classification problem.}
\scalebox{1}{ \!\!\!\setlength{\tabcolsep}{3.5pt}
\begin{tabular}{lllllc}    	\addlinespace
    \toprule
Layer&$D\times W \times H$& $K\times K$ &$P$&$S$&$\sigma$\\[0pt]   
\cmidrule{1-6}
\multirow{1}{*}{Input}&$1\times28\times28$&-&-&-&-\\[2pt]
Convolutional&$32\times27\times27$ &$4\times4$&$1$&$1$&ReLU\\[2pt]
Pooling&$32\times13\times13$ &$3\times3$&$0$&$2$&-\\[2pt]
Convolutional&$64\times9\times9$ &$5\times5$&$0$&$1$&ReLU\\[2pt]
Pooling&$64\times4\times4$ &$3\times3$&$0$&$2$&-\\[2pt]
Fully connected&$150\times1\times1$ &-&-&-&ReLU\\[2pt]
Output&$11\times1\times1$ &-&-&-&-\\[2pt]
\bottomrule
\end{tabular}}
\end{center}
\end{table}

\subsubsection{CIFAR10}\label{C:cifar10)}
\begin{table}[h!]
\begin{center}
\caption{Configuration details for the CNN applied to the CIFAR10 classification problem.}
\scalebox{1}{ \!\!\!\setlength{\tabcolsep}{3.5pt}
\begin{tabular}{lllllc}    	\addlinespace
    \toprule
Layer&$D\times W \times H$& $K\times K$ &$P$&$S$&$\sigma$\\[0pt]   
\cmidrule{1-6}
\multirow{1}{*}{Input}&$3\times32\times32$&-&-&-&-\\[2pt]
Convolutional&$32\times32\times32$ &$5\times5$&$2$&$1$&ReLU\\[2pt]
Pooling&$32\times16\times16$ &$3\times3$&$1$&$2$&-\\[2pt]
Convolutional&$32\times16\times16$ &$5\times5$&$2$&$1$&ReLU\\[2pt]
Average pooling&$32\times8\times8$ &$3\times3$&$1$&$2$&-\\[2pt]
Convolutional&$64\times8\times8$ &$5\times5$&$2$&$1$&ReLU\\[2pt]
Average pooling&$64\times4\times4$ &$3\times3$&$1$&$2$&-\\[2pt]
Fully connected&$64\times1\times1$ &-&-&-&ReLU\\[2pt]
Output&$11\times1\times1$ &-&-&-&-\\[2pt]
\bottomrule
\end{tabular}}
\end{center}
\end{table}
\newpage
\subsection{infoGAN}\label{B:GAN}

\subsubsection{MNIST}\label{BG:mnist}
\begin{table}[h!]
\begin{center}
\caption{Configuration details for Dnet in the experiment on the MNIST Dataset. The leaky rate for lReLU is set to 0.2.}
\scalebox{1}{ \!\!\!\setlength{\tabcolsep}{3.5pt}
\begin{tabular}{lllllc}    	\addlinespace
    \toprule
Layer&$D\times W \times H$& $K\times K$ &$P$&$S$&$\sigma$\\[0pt]   
\cmidrule{1-6}
\multirow{1}{*}{Input}&$1\times28\times28$&-&-&-&-\\[2pt]
Convolutional&$32\times28\times28$ &$3\times3$&$1$&$1$&lReLU\\[2pt]
Batch normalization&$32\times28\times28$ &-&-&-&\\[2pt]
Average pooling&$32\times14\times14$ &$3\times3$&$1$&$2$&-\\[2pt]
Convolutional&$64\times14\times14$ &$3\times3$&$1$&$1$&lReLU\\[2pt]
Batch normalization&$64\times14\times14$ &-&-&-&\\[2pt]
Average pooling&$64\times7\times7$ &$3\times3$&$1$&$2$&-\\[2pt]
Output&$512\times1\times1$ &-&-&-&lReLU\\[2pt]
\bottomrule
\end{tabular}}
\end{center}
\end{table}

\begin{table}[h!]
\begin{center}
\caption{Configuration details for Pnet in the experiment on the MNIST Dataset.}
\scalebox{1}{ \!\!\!\setlength{\tabcolsep}{3.5pt}
\begin{tabular}{lllllc}    	\addlinespace
    \toprule
Layer&$D\times W \times H$& $K\times K$ &$P$&$S$&$\sigma$\\[0pt]   
\cmidrule{1-6}
\multirow{1}{*}{Input}&$512\times1\times1$&-&-&-&-\\[2pt]
Output&$1\times1\times1$ &-&-&-&-\\[2pt]
\bottomrule
\end{tabular}}
\end{center}
\end{table}

\begin{table}[h!]
\begin{center}
\caption{Configuration details for Qnet in the experiment on the MNIST Dataset.}
\scalebox{1}{ \!\!\!\setlength{\tabcolsep}{3.5pt}
\begin{tabular}{lllllc}    	\addlinespace
    \toprule
Layer&$D\times W \times H$& $K\times K$ &$P$&$S$&$\sigma$\\[0pt]   
\cmidrule{1-6}
\multirow{1}{*}{Input}&$512\times1\times1$&-&-&-&-\\[2pt]
Fully connected&$300\times1\times1$ &-&-&-&ReLU\\[2pt]
Output&$13\times1\times1$ &-&-&-&-\\[2pt]
\bottomrule
\end{tabular}}
\end{center}
\end{table}

\begin{table}[h!]
\begin{center}
\caption{Configuration details for Gnet in the experiment on the MNIST Dataset.}
\scalebox{1}{ \!\!\!\setlength{\tabcolsep}{3.5pt}
\begin{tabular}{lllllc}    	\addlinespace
    \toprule
Layer&$D\times W \times H$& $K\times K$ &$P$&$S$&$\sigma$\\[0pt]   
\cmidrule{1-6}
\multirow{1}{*}{Input}&$75\times1\times1$&-&-&-&-\\[2pt]
Fully connected&$3072\times1\times1$ &-&-&-&ReLU\\[2pt]
Transposed convolutional&$64\times7\times7$ &$3\times3$&$1$&$1$&ReLU\\[2pt]
Transposed convolutional&$32\times14\times14$ &$3\times3$&$1$&$2$&ReLU\\[2pt]
Output&$1\times28\times28$ &$3\times3$&$1$&$2$&-\\[2pt]
\bottomrule
\end{tabular}}
\end{center}
\end{table}
\newpage
\subsubsection{CelebA}\label{BG:celebA}
\begin{table}[h!]
\begin{center}
\caption{Configuration details for Dnet in the experiment on the CelebA Dataset. The leaky rate for lReLU is set to 0.2.}
\scalebox{1}{ \!\!\!\setlength{\tabcolsep}{3.5pt}
\begin{tabular}{lllllc}    	\addlinespace
    \toprule
Layer&$D\times W \times H$& $K\times K$ &$P$&$S$&$\sigma$\\[0pt]   
\cmidrule{1-6}
\multirow{1}{*}{Input}&$3\times32\times32$&-&-&-&-\\[2pt]
Convolutional&$32\times32\times32$ &$3\times3$&$1$&$1$&lReLU\\[2pt]
Batch normalization&$32\times32\times32$ &-&-&-&\\[2pt]
Average pooling&$32\times16\times16$ &$3\times3$&$1$&$2$&-\\[2pt]
Convolutional&$32\times16\times16$ &$3\times3$&$1$&$1$&lReLU\\[2pt]
Batch normalization&$32\times16\times16$ &-&-&-&\\[2pt]
Average pooling&$32\times8\times8$ &$3\times3$&$1$&$2$&-\\[2pt]
Convolutional&$64\times8\times8$ &$3\times3$&$1$&$1$&lReLU\\[2pt]
Batch normalization&$64\times8\times8$ &-&-&-&\\[2pt]
Average pooling&$64\times4\times4$ &$3\times3$&$1$&$2$&-\\[2pt]
Output&$256\times1\times1$ &-&-&-&lReLU\\[2pt]
\bottomrule
\end{tabular}}
\end{center}
\end{table}
\newpage

\begin{table}[h!]
\begin{center}
\caption{Configuration details for Pnet in the experiment on the CelebA Dataset.}
\scalebox{1}{ \!\!\!\setlength{\tabcolsep}{3.5pt}
\begin{tabular}{lllllc}    	\addlinespace
    \toprule
Layer&$D\times W \times H$& $K\times K$ &$P$&$S$&$\sigma$\\[0pt]   
\cmidrule{1-6}
\multirow{1}{*}{Input}&$256\times1\times1$&-&-&-&-\\[2pt]
Output&$1\times1\times1$ &-&-&-&-\\[2pt]
\bottomrule
\end{tabular}}
\end{center}
\end{table}
\begin{table}[h!]
\begin{center}
\caption{Configuration details for Qnet in the experiment on the CelebA Dataset.}
\scalebox{1}{ \!\!\!\setlength{\tabcolsep}{3.5pt}
\begin{tabular}{lllllc}    	\addlinespace
    \toprule
Layer&$D\times W \times H$& $K\times K$ &$P$&$S$&$\sigma$\\[0pt]   
\cmidrule{1-6}
\multirow{1}{*}{Input}&$256\times1\times1$&-&-&-&-\\[2pt]
Fully connected&$256\times1\times1$ &-&-&-&ReLU\\[2pt]
Output&$110\times1\times1$ &-&-&-&-\\[2pt]
\bottomrule
\end{tabular}}
\end{center}
\end{table}

\begin{table}[h!]
\begin{center}
\caption{Configuration details for Gnet in the experiment on the CelebA Dataset.}
\scalebox{1}{ \!\!\!\setlength{\tabcolsep}{3.5pt}
\begin{tabular}{lllllc}    	\addlinespace
    \toprule
Layer&$D\times W \times H$& $K\times K$ &$P$&$S$&$\sigma$\\[0pt]   
\cmidrule{1-6}
\multirow{1}{*}{Input}&$238\times1\times1$&-&-&-&-\\[2pt]
Fully connected&$1024\times1\times1$ &-&-&-&ReLU\\[2pt]
Transposed convolutional&$64\times4\times4$ &$3\times3$&$1$&$1$&ReLU\\[2pt]
Transposed convolutional&$64\times8\times8$ &$3\times3$&$1$&$2$&ReLU\\[2pt]
Transposed convolutional&$32\times16\times16$ &$3\times3$&$1$&$2$&ReLU\\[2pt]
Transposed convolutional&$32\times32\times32$ &$3\times3$&$1$&$2$&ReLU\\[2pt]
Transposed convolutional&$32\times32\times32$ &$3\times3$&$1$&$1$&ReLU\\[2pt]
Output&$3\times32\times32$ &$3\times3$&$1$&$2$&-\\[2pt]
\bottomrule
\end{tabular}}
\end{center}
\end{table}
\newpage
\section{CelebA latent space variations}\label{Appendix:C}\vspace{-5mm}
\begin{figure}[h!]\centering
\subfloat[Backpropagation - Gender]{\includegraphics[width=60mm]{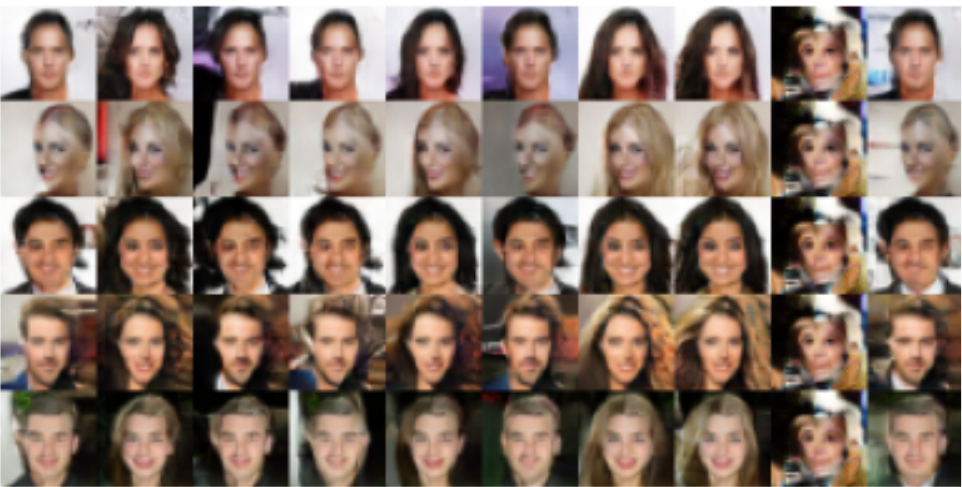}}
\hspace{0.3cm}\subfloat[TAGI -- Gender]{\includegraphics[width=60mm]{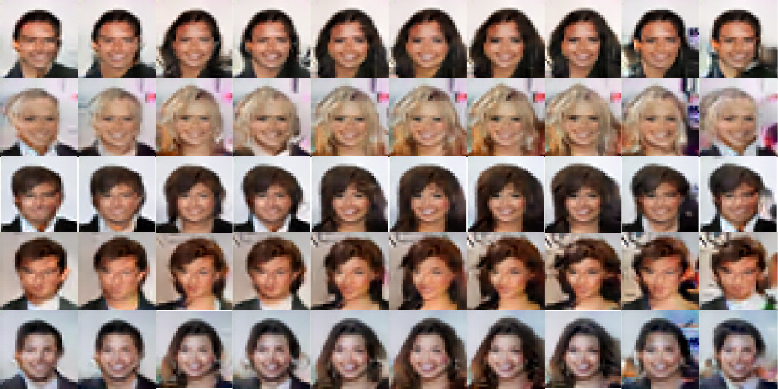}}\\
\subfloat[Backpropagation -- Azimuth]{\includegraphics[width=60mm]{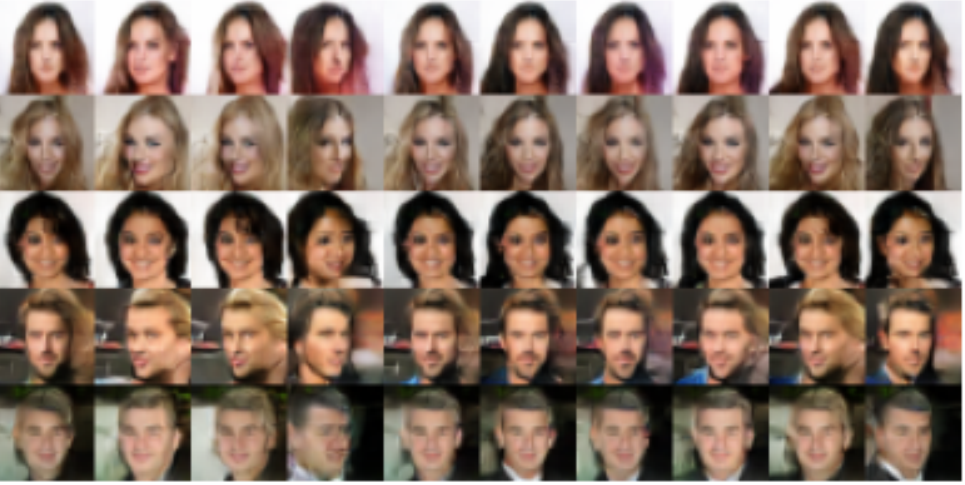}}
\hspace{0.3cm}\subfloat[TAGI -- Azimuth]{\includegraphics[width=60mm]{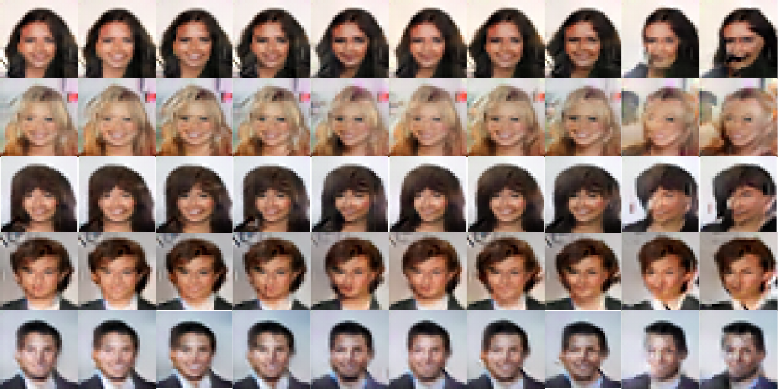}}\\
\subfloat[Backpropagation -- Hair style]{\includegraphics[width=60mm]{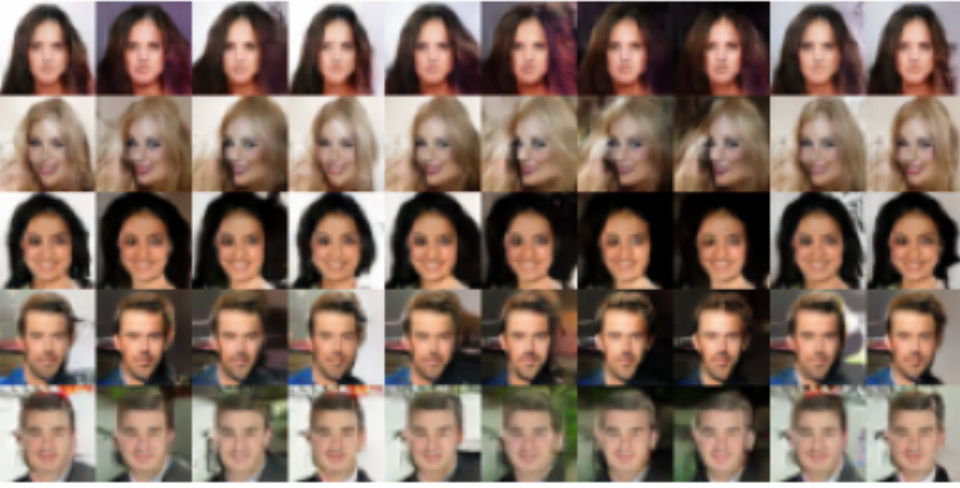}}
\hspace{0.3cm}\subfloat[TAGI -- Hair style]{\includegraphics[width=60mm]{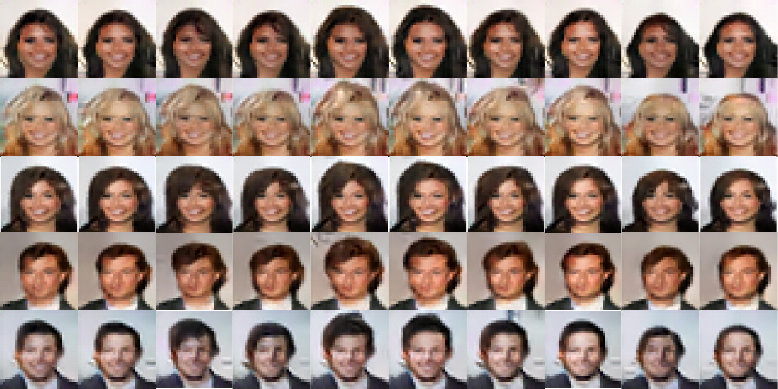}}\\
\subfloat[Backpropagation -- contrast]{\includegraphics[width=60mm]{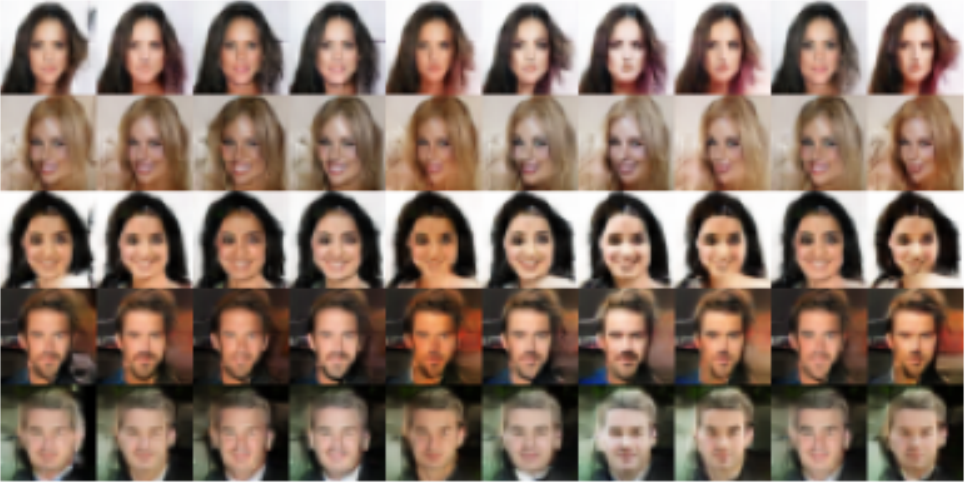}}
\hspace{0.3cm}\subfloat[TAGI -- constrast]{\includegraphics[width=60mm]{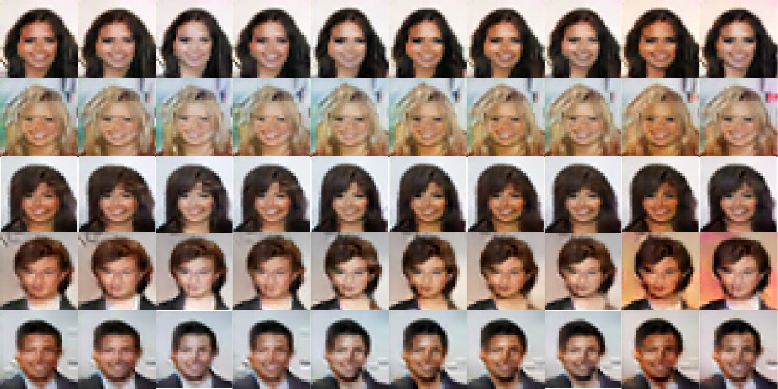}}
\caption{Comparison of the capacity to control the latent space for the architecture presented in \cite{chen2016infogan} and trained with backpropagation, and a network six times smaller trained with TAGI over half as many epochs.}
\end{figure}
\section{Supplementary Results for Classification}\label{Appendix:D}
\begin{table}[h!]
\centering
\caption{Comparison of the classification performance on CIFAR-10  between a network trained with TAGI and backpropagation (BP), BP, MC-dropout, and VOGN on residual networks over three runs. ($\mathtt{E}$: number of epochs; $\mathtt{B}$: batch size ; DA: data augmentation).}
\scalebox{0.9}{\begin{tabular}{cccc|cc}    	\addlinespace
    \toprule
    	\multicolumn{1}{l}{}&\multicolumn{1}{l}{}&\multicolumn{2}{c}{Error rate [\%]}&\multicolumn{2}{|c}{Hyperparameters}\\[0pt] 
	\cmidrule{2-6}
Architecture&Method&$e=1$& $e=\mathtt{E}$ &$\mathtt{E}$&$\mathtt{B}$\\[0pt]   
\cmidrule{1-6}
\multirow{4.5}{*}{Resnet18}&\bf TAGI&$65.9\pm0.667$ &$13.8\pm0.921$ &$50$ &16\\[2pt]
\multirow{4.5}{*}{(DA)}&BP~\cite{osawa2019practical}&- &$14.0\pm0.257$ &$160$ &$128$\\[2pt]
&MC-dropout~\cite{osawa2019practical}&- &$17.2\pm0.208$ &$161$ &$128$\\[2pt]
&VOGN~\cite{osawa2019practical}&- &$15.7\pm0.195$ &$161$ &$128$\\[2pt]
\cmidrule{1-6}
\multirow{1}{*}{\begin{tabular}{c}Resnet18half\\[2pt](DA)\end{tabular}}&& &&&\\[-4pt]
\multirow{-1}{*}{}&\bf TAGI&$65.9\pm0.4$ &$14.9\pm0.7$ &$50$ &16\\[6pt]
\bottomrule
\end{tabular}}
\label{TAB:CIFAR10_std_cls}
\end{table}

\begin{table}[h!]
\centering
\caption{Comparison of the predictive uncertainty calibration performance on CIFAR-10 between a network trained with TAGI and backpropagation (BP), BP, MC-dropout, and VOGN over three runs. ($\mathtt{E}$: number of epochs; $\mathtt{B}$: batch size ; NLL: Negative Log-Likelihood; ECE: Expected Calibration Error; AUROC: Area Under the Receiver Operating Characteristics).}
\scalebox{0.75}{\begin{tabular}{ccccc|cc}    	\addlinespace
    \toprule
    	\multicolumn{2}{l}{}&\multicolumn{1}{c}{NLL}&\multicolumn{1}{c}{ECE}&\multicolumn{1}{c}{AUROC}&\multicolumn{2}{|c}{Hyperparameters}\\[0pt] 
	\cmidrule{2-7}
Architecture&Method&$e=\mathtt{E}$&$e=\mathtt{E}$ &$e=\mathtt{E}$&$\mathtt{E}$&$\mathtt{B}$\\[0pt]   
\cmidrule{1-7}
\multirow{4.5}{*}{Resnet18}&\bf TAGI &0.588 $\pm$ 0.09& 0.143$\pm$0.08&0.98$\pm$0.001&$50$ &16\\[2pt]
\multirow{4.5}{*}{(DA)}&BP~\cite{osawa2019practical}&0.55$\pm$0.01&0.082$\pm$0.002& 0.877$\pm$0.001&$160$ &$128$\\[2pt]
&MC-dropout~\cite{osawa2019practical} &0.51$\pm$0&0.166$\pm$0.025& 0.768$\pm$0.004&$161$ &$128$\\[2pt]
&VOGN~\cite{osawa2019practical} &0.477$\pm$0.006&0.04$\pm$0.002& 0.876 $\pm$0.002&$161$ &$128$\\[2pt]
\cmidrule{1-7}
\multirow{1}{*}{\begin{tabular}{c}Resnet18half\\[2pt](DA)\end{tabular}}&\bf && &&$$ &\\[-4pt]
&\bf TAGI &0.541$\pm$0.04& 0.046$\pm$0.001&0.978$\pm$0.003&$50$ &16\\[6pt]
\bottomrule
\end{tabular}}
\label{TAB:CIFAR10_std_cali}
\end{table}

\end{document}

%% file: tikz_setup_GAN.tex
\definecolor{babyblue}{rgb}{0.54, 0.81, 0.94}
\definecolor{light-gray}{gray}{0.85}
\definecolor{amber}{rgb}{1.0, 0.49, 0.0}
\tikzstyle{connect}=[-latex, thick,>=latex]
\tikzstyle{shaded}=[draw=black!20,fill=black!20]
\tikzstyle{line} = [draw, -latex,thick,>=latex]

\tikzstyle{input}=[draw,fill=red!10,rectangle,minimum width=1cm, minimum height = 20pt, inner sep=0pt, rounded corners = 2pt]
\tikzstyle{hidden}=[draw,fill=green!20,circle,minimum size=20pt,inner sep=0pt]
\tikzstyle{conv}=[draw,fill=green!20,rectangle, minimum width=3cm, minimum height = 20pt,inner sep=0pt, rounded corners = 2pt]
\tikzstyle{output}=[draw,fill=babyblue!20,rectangle,minimum width=1.5cm, minimum height = 20pt, inner sep=0pt, rounded corners = 2pt]
\tikzstyle{output2}=[draw,fill=babyblue!20,rectangle,minimum width=2.2cm, minimum height = 20pt, inner sep=0pt, rounded corners = 2pt]
\tikzstyle{bias}=[draw=none,circle,minimum size=20pt,inner sep=0pt]
\tikzstyle{biasB}=[draw=none,circle,minimum size=28pt,inner sep=0pt]
\tikzstyle{activ}=[draw,fill=cyan!20,circle,minimum size=20pt,inner sep=0pt]
\tikzstyle{activConv}=[draw,fill=cyan!20,rectangle, minimum size=20pt,inner sep=0pt, rounded corners = 2pt]
\tikzstyle{norm}=[draw,fill=amber!30,rectangle,minimum width=3cm, minimum height = 20pt,inner sep=0pt, rounded corners = 2pt]
\tikzstyle{pool}=[draw,fill=yellow!30,rectangle,minimum width=3cm, minimum height = 20pt,inner sep=0pt, rounded corners = 2pt]
\tikzstyle{obs}=[draw,fill=blue!10,rectangle,minimum width=1.5cm, minimum height = 20pt, inner sep=0pt, rounded corners = 2pt]

\tikzstyle{stateTransition}=[->,line width=0.25pt,draw=black!50]
\tikzstyle{Forward}=[->,line width=0.25pt,draw=magenta!75]
\tikzstyle{Inference}=[->,line width=0.25pt,draw=cyan!75]
\tikzstyle{dnet}=[draw,fill=green!20,rectangle, minimum width=1cm, minimum height = 20pt,inner sep=0pt, rounded corners = 2pt]
\tikzstyle{gnet}=[draw,fill=babyblue!20,rectangle, minimum width=1cm, minimum height = 20pt,inner sep=0pt, rounded corners = 2pt]
\tikzstyle{disc}=[ellipse, minimum height = 1.2 cm, minimum width=1.2cm, thick, draw =black!80,inner sep=0pt, fill=green!20]
\tikzstyle{discD}=[ellipse, minimum height = 1.2 cm, minimum width=1.2cm, thick, draw =white!80,inner sep=0pt]
\tikzstyle{dummybox}=[rectangle, minimum width=2.2cm,minimum height=1.2cm,draw=white!100, thin,rounded corners]
\tikzstyle{normLayer}=[rectangle, fill=amber!30, minimum width=18pt,minimum height=4.8cm,draw=black!100, thin,rounded corners]